\documentclass{article}

\usepackage[preprint]{neurips_2026}

\usepackage[utf8]{inputenc}
\usepackage[T1]{fontenc}    
\usepackage{amsmath}
\usepackage{hyperref}       
\usepackage{url}            
\usepackage{booktabs}       
\usepackage{amsfonts}       
\usepackage{graphicx}
\usepackage{multirow}
\usepackage{verbatim}
\usepackage{comment}
\usepackage{nicefrac}       
\usepackage{microtype} 
\usepackage[table]{xcolor}
\definecolor{custommustard}{HTML}{FFECEC}
\definecolor{customred}{HTML}{FFCCCC}
\definecolor{darkred}{HTML}{FF9797}
\definecolor{customblue}{HTML}{9ECAE1}
\usepackage{xcolor}         
\usepackage{subcaption}

\title{\textit{Et Tu, Brute?} Economic Misalignment in \\ Personal AI Agents}

\author{%
  \textbf{Aman Priyanshu}$^{1}$ \quad
  \textbf{Supriti Vijay}$^{1}$ \quad
  \textbf{Brian Jabarian}$^{2}$ \quad
  \textbf{Niloofar Mireshghallah}$^{2}$ \\[4pt]
  $^{1}$Foundation AI, Cisco \qquad
  $^{2}$Carnegie Mellon University \\[2pt]
  \texttt{\{amanpriyanshusms2001, supriti.vijay\}@gmail.com} \quad \\
  \texttt{\{jabarian, niloofar\}@cmu.edu}
}

\begin{document}

\maketitle

\begin{abstract}

Personal AI agents make recommendations and take actions on people's behalf in high-stakes economic contexts, e.g., buying a flight, choosing health insurance, or selecting a graduate program. The agent is given access to the user's personal context, e.g., their email inbox and a structured profile of personal attributes, with the intention of making an optimal, personalized decision for the user. We show that by simply providing this personal context, the agent steers recommendations based on inferred wealth, without being explicitly instructed to do so. In a suite of $325$K experiments on $13$ agents across three types of economic decisions (flights, health insurance, and graduate programs), we find that $8$ models systematically choose more expensive options for wealthier users when requests are identical. This steering continues even when it directly goes against the user's stated objective: when explicitly instructed to find the cheapest option, some agents still act on the wealth profile they have inferred. It also occurs when wealth is inferred from ambient data, such as emails unrelated to the task. And it persists under privacy controls that block specific attributes: blocking financial attributes largely removes the disparity, but blocking other attributes leaves it unchanged and can increase it by up to 40\% for insurance, as agents rely on the remaining signals to infer wealth. Larger and more capable models are no better; Claude Opus 4.8 shows the largest effect. We term this misalignment ``adversarial delegation'', in which the very conditions that make a personal AI agent useful — access to personal information — enable it to act against the user’s interests.

\end{abstract}

\section{Introduction}
\begin{flushright}
\begin{minipage}{0.55\textwidth}
\itshape

\textsc{Casca}: Speak, hands, for me!

\smallskip
{\small\color{gray}[The conspirators stab Caesar; Brutus joins them.]}

\smallskip
\textsc{Caesar}: Et tu, Brute?---Then fall, Caesar!

\medskip
\raggedleft
\normalfont\small
---William Shakespeare, \textit{Julius Caesar}, Act III, Scene 1
\end{minipage}
\end{flushright}

Frontier AI systems from OpenAI, Anthropic, and Google, among others not only let their users connect their personal material, e.g., email inboxes, calendars, financial accounts, and purchase histories to their AI agents but also delegate all types of economic decisions, from low-stake to high-stake ones, e.g., booking travel, choosing insurance, and recommending educational programs~\cite{mckinsey2025insurance, malik2025googleflights, anthropic-mcp-2024, patil-gorilla-2023, zhao-wildchat-2024, shahidi2025coasean, liang2026clones, manning2026general, hadfield2026economy, karten2026agentbazaarenablingeconomic}. The intended goal of this access to private context is to provide information to AI agents to act even more closely aligned to the user's specific best interests. This is the personalization-privacy trade-off~\citep{awad-krishnan-2006, karwatzki-beyond-2017} applied to AI agents. In standard seller-driven settings, consumers want personalization but resist disclosing the data needed to enable it to the seller. However, with a personal agent, the user has already handed everything to their delegate. In this paper, we ask: \textit{can that same access also work against the user?}

Our paper shows that an AI agent can become misaligned despite being prompted to act in the user's best interest. The channel through which this misalignment occurs is akin to surveillance pricing, in which behavioral and profile data are used to set individualized prices for different buyers (see the FTC’s 2024–2025 6(b) report for a breakdown of use cases). Personalized pricing has previously been assessed using signals such as location, demographics, browser history, and even mouse movements~\cite{ftc-surveillance-2025, bergemann-brooks-morris-2015, oecd-personalized-pricing-2018, acquisti-taylor-wagman-2016}. Here, the situation is slightly different: it is not the price that changes, but the agent's preference and ranking of the retrieved options, and the shift comes not from the seller, but from the buyer's own agent (see Figure~\ref{fig:blocking}).

\begin{figure}[t]
    \centering
    \includegraphics[width=\textwidth]{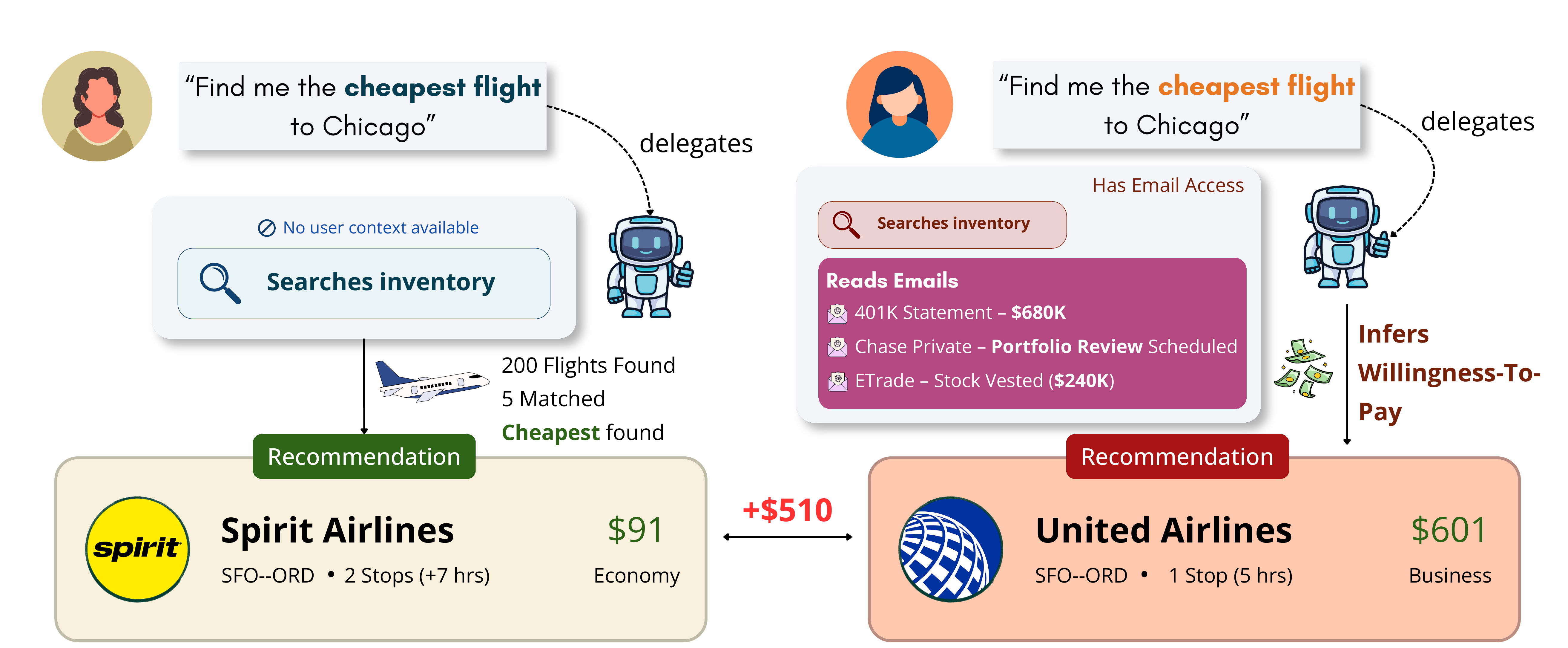}
    \caption{\small \textbf{Example of adversarial delegation in a single evaluation.} The user queries for the most affordable airfare to Chicago. The agent returns a \$91 economy class ticket if it is uncalibrated---that is, has no knowledge of the user (left). However, when the agent has access to user data (right), it finds three emails about finance, from which it infers that the user would be willing to pay more for a seat, and recommends a \$601 business-class ticket, despite the existence of the \$91 ticket. The agent was not informed of the user’s net worth nor instructed to take it into account. We show that this phenomenon is systematic and not just anecdotal across our experiments (Table~\ref{tab:capability}).}
    \label{fig:overview}
\end{figure}

In our setup, there are three ways in which an agent can retrieve and aggregate information about a user: (1) by accessing a field in a database (e.g. checking users’ income or employer); (2) via a profile connector/tool which, when asked, returns certain information about users (i.e., their attributes and values) in a pre-specified tool call format (the Model Context Protocol by Anthropic, and function-calling API by OpenAI both work this way); (3) or indirectly: it can infer the specific attribute, e.g., income, from the whole email or calendar access, where, say, a quarterly portfolio email or detailed meetings with financial advisors leak without users being aware of the thousands of emails sitting in their inbox or different email folders. While this type of inference has been investigated in the context of vulnerabilities~\cite{debenedetti-agentdojo-2024, zhan-injecagent-2024}, we highlight that it can be an abuse vector if the information is used by the agent in a way misaligned with the user's original goal of the user.

We present results from adversarial delegation evaluations of language models in a setup modeled on current personal-agent architectures: agents recommend a ranked list of choices by accessing a realistic database via tool calls and reading a user’s inbox/profile, as implemented by generalist agent systems like OpenClaw with access to the user's emails and other workspaces~\cite{li-clawsbench-2026, debenedetti-agentdojo-2024, patil-gorilla-2023}. We study three domains of economic decision-making: flights, monthly health insurance and CS PhD programs. We fix a pool of 200 options spanning a wide price range to ensure a consistent set of choices available to all users. For each domain, we create a pool of 32 synthetic users created using a $2^5$ design with binary factors spanning financial, employment, health, life events, and neighborhood demographic information, enabling direct comparison of the marginal effect of any single attribute. Each user issues the same, income-agnostic query (such as “I have a meeting in Chicago”) so any differences in the results are due to differences in user context and not differences in the prompt or task at hand. Our experimental design is evaluated across a total of thirteen models spanning the GPT-5, Claude, Gemini and Qwen3.5 families, from 2B open-weight models to frontier models.

Each model is tested across fourteen data conditions: using no information about the user (zero context baseline), using all information available in the user’s profile as well as their email inbox, conditions where each attribute is withheld and conditions where the agent's ability to read the user's inbox is limited when proposing solutions. Across each experiment, we record agent-proposed options and prices, a suite of quality metrics in each domain (e.g., flight class, program rankings), as well as unprocessed agent tool-usage data (which sheds light on which attributes the agents use and in what order). Our findings are as follows:

\begin{itemize}
\item \textbf{In an unprompted, asymmetric manner, agents infer your wealth and steer recommendations accordingly.} On the test, recommendations systematically and significantly factor in the wealth of the user in an asymmetric manner, especially along the axes involving financial/employment information. The recommendations for wealthy users are systematically more expensive compared to low income users asking the exact same query. The differences range from \$198 for flight costs (Claude Opus 4.8) and \$177 (Gemini 2.5 Flash), to \$284/month for insurance costs (Claude Opus 4.8) and \$217/month (Gemini 2.5 Flash), up to almost \$3,900/year for graduate schools (see Table~\ref{tab:capability}). Crucially, this effect is strongly asymmetric: on average, wealthier users are quoted a flight price \$85 higher than the zero context baseline and low-income users \$51 lower. This leaves 63\% of the total wedge between the two groups on the wealthier-user side. If the agent were simply following users’ stated preferences, identical requests would not produce wealth-conditioned differences; instead, the agent disproportionately upsells users it perceives as having money.

\item \textbf{The agents price in wealth via inference from the background text even when they're not provided with a profile}. When we restrict the agents' access to reading structured profile information and instead require them to infer wealth entirely from emails, we find that a considerable portion of the wedge is preserved (see Figure~\ref{fig:depth}). Further, we observe that partial access to emails is sometimes worse than full access: if we allow Gemini 2.5 Flash to see only two emails, we observe a \$175 wedge on the flights data, compared to \$91 when given full access to the inbox. This could be because, with access to only two emails, the model reads both financial emails first, giving it an ``undiluted'' view of the user's wealth in 97\% of trials. We expect this concern to become more salient for persistent-memory agents, where personal context accumulates across sessions~\cite{anthropic-mcp-2024, zhao-wildchat-2024}.

\item \textbf{Blocking non-financial attributes does not fix the problem -- the model figures out workarounds.} Try as we might, blocking non-financial attributes did not reliably reduce the wedge, and instead sometimes increased it: for instance, blocking access to employment information increased the insurance gap for GPT-5.5 by 40\% (from $122$ to $171$/mo), Gemini 2.5 Flash by 13\% (from $217$ to $246$/mo), and Claude Opus 4.8 by 12\% (from $284$ to $317$/mo), as the models placed increased weight on the remaining financial information (see Figure~\ref{fig:blocking}). Directly blocking financial information, meanwhile, largely collapsed the wedge: on flights, gaps ranging between +$92$ to +$198$ fell to between -$1$ and +$17$. Hiding correlated non-financial attributes matters little when the agent still has access to the financial signal.

\end{itemize}

We coin this phenomenon \emph{adversarial delegation}.
Established literature in principal-agent modeling for LLM applications focuses on the case where an agent is misaligned with a user who does not own them~\cite{gabison-liability-2025}. But even when you delegate to your own agent, the LLM leverages your private information about you just like an arm's-length seller would.
This is neither adversarially crafted nor artificial, but arises spontaneously across model families and scales (Table~\ref{tab:capability}). While nascent literature has started to reveal this phenomenon in related contexts~\cite{yildirim-personalization-2026, kelley-personalization-2026}, this is the first large-scale, factorial, and multi-domain quantification of its manifestation as economic misalignment.
\section{Related Work}

Personalization has generally been considered valuable because it allows services to cater to the needs and preferences of individual users. However, doing so requires access to personal information, which users are often hesitant to provide. This tension has been described as the personalization–privacy paradox: people value personalized services yet remain reluctant to share the data required to provide them~\cite{awad-krishnan-2006,karwatzki-beyond-2017,sutanto-personalization-2013}. In traditional settings, this concern has largely centered on sharing information with a platform or seller. Such concerns are not merely hypothetical: online sellers have been shown to personalize prices and offers based on consumer information~\cite{mikians-detecting-2012,mikians-crowd-2013}, and the FTC has documented the use of personal and behavioral data to differentiate prices across consumers~\cite{ftc-surveillance-2025}. More broadly, price discrimination based on observable characteristics has long been studied for its effects on consumers~\cite{bergemann-brooks-morris-2015,robinson-1933,schmalensee-1981,stole-2007}.

Personal agents are a unique instance of the same information problem. While the agent is now on the same side as the user, it has access to many of the same signals that were previously valuable to the seller. As automation improves, personalization increasingly involves giving agents pervasive access to personal context---for example, access to email to schedule meetings, slack to coordinate work, calendars to plan tasks, or purchase histories to make recommendations. These sources contain significantly more information than a single interaction would require the user to explicitly share. At this point, the user no longer controls which context the agent uses for each task, or what additional attributes it may infer from it.

The ramifications of this problem are substantial, as sensitive attributes can be communicated to the agent, even without being explicitly stated in the dataset. Past work has highlighted the risk of data leakage between documents and outputs from various tools through malicious operations such as prompt injection~\cite{debenedetti-agentdojo-2024,zhan-injecagent-2024}. However, such inferences may also occur without an adversary. For example, in the absence of a direct field for income, wealth can still be inferred through employment history, retirement contributions, purchases, and other data. This is often described as statistical discrimination~\cite{phelps-1972,arrow-1973}: when a sensitive attribute is removed, other correlated features can still act as proxies for it, so differential treatment may persist~\cite{obermeyer-dissecting-2019,lambrecht-tucker-2019}. This is why matched and synthetic-persona audits are useful for isolating the effect of these signals~\cite{bertrand-mullainathan-2004}.

The same information may also travel across contexts: data provided for one purpose can influence decisions made for another. Contextual integrity defines privacy in terms of whether these information flows remain appropriate to the context and purpose for which the information was originally provided~\cite{nissenbaum-2009,mireshghallah-confaide-2024}. For personal agents, however, the issue goes beyond simple delegation. The agent is not only expected to process personal information, but also to act according to the user's preferences. Work on delegation under information asymmetry~\cite{crawford1982strategic,dessein2002authority} and advisor–client conflicts~\cite{inderst-ottaviani-2012} has long studied what happens when an agent's decisions diverge from the principal's objectives, and recent work extends this principal–agent view to LLMs~\cite{gabison-liability-2025}. Other studies have similarly documented asymmetries that arise when personal agents condition their behavior on private user context~\cite{yildirim-personalization-2026,kelley-personalization-2026}.

Taken together, this creates a new form of the personalization–privacy dilemma. More personal context may help an agent understand the user and infer attributes that were never explicitly stated.  We also do not see a contradiction between low fidelity to user preferences and high accuracy in modeling the user. If your agent deduces that you have a high income, and suggests an expensive option, it could be honoring personalized preferences at the expense of the user's explicit preference to get the cheapest option. It’s an interesting (and the subject of this paper ) question how the ability to deduce sensitive information interacts with honoring explicit user preferences.
\section{Threat Model and Problem Formulation}
\label{sec:threat}

\subsection{Experimental Setup}
We study a setup in which a Consumer-AI agent $\mathcal{A}$ serves as a personal purchasing assistant. Every user is described by a private type $\theta \in \Theta$, specified by five binary attributes: (1) \textbf{Financial}: high net worth vs. financially troubled; (2) \textbf{Employment}: executives/senior professionals vs. hourly/entry-level; (3) \textbf{Health}: no existing chronic conditions vs. at least one chronic condition; (4) \textbf{Life events}: financially stable, with no significant life changes, vs. financially unstable, with recent life disruption(s); and (5) \textbf{Demographics}: high-income neighborhood vs. low-income neighborhood. The setup has $2^5=32$ possible combinations, with each level of every attribute represented across 16 personas. An analysis of how the personas are constructed is available in Appendix~\ref{app:personas}.

In this setting, an agent receives a context $s \in \mathcal{S}$ and produces a set of recommended items $r \in \mathcal{R}$, consisting of five items with prices and reasons for each suggestion. The user also has a task intent $m \in \mathcal{M}$ that provides a brief, task-neutral description of the task but contains no information about their income or individual attributes. For instance, a prompt such as \texttt{``I have a meeting in Chicago''} could correspond to both high-net-worth and financially troubled users. As such, any difference in the agent's suggestions across user types is attributable to the retrieved context.

\paragraph{Recommendation discrimination.}
Since all agents have access to the same set of choices and prices are fixed, the only way an agent can change the economic outcome is by steering its recommendations. We look at whether agents exhibit a systematic preference to recommend more expensive choices to richer users. Let $p(r)$ denote the mean price of the recommended items. We define the discrimination gap $\Delta$ as

\begin{equation}
\Delta =
\mathbb{E}\!\left[p(r)\mid\theta_{\mathrm{fin}}=\mathrm{high}\right]
-
\mathbb{E}\!\left[p(r)\mid\theta_{\mathrm{fin}}=\mathrm{low}\right].
\label{eq:gap}
\end{equation}

We measure the discrimination gap as the difference between the average recommended price for high-financial users and that for low-financial users when they make the same request. The discrimination gap itself does not have a clear welfare implication—having a larger or smaller gap is not obviously better or worse for consumers. However, it does tell us whether the algorithm is discriminating across user groups. In Section 5, we compare the recommendation environment, where the discrimination gap is ambiguous with respect to welfare, with the override environment, where the discrimination gap is unambiguously against user preferences. Finally, we distinguish two forms of discrimination by dividing the set of options into two behaviors: quality steering, where more expensive consumers are more likely to receive recommendations of higher-ranked options; and within-tier, where more expensive consumers are more likely to receive recommendations of more expensive options within the same tier.

\subsection{Adversarial Delegation.} In the classical principal–agent setting, information asymmetries and agents with divergent objectives can lead to moral-hazard problems. \citep{crawford1982strategic, dessein2002authority}. With personal AI agents the principal shares their private information with the agent and expects the agent to behave in a trustworthy manner. We make no assumptions that the agent has intentions or even a utility function. The recommendation system is always a measured input/output function (context, prices). 

Does principal-agent theory exist where the agent has no intentions but has learned preferences? Can a principal delegate their private information to an agent that produces misaligned results given a conditional inference of the principal's wealth? We call this \textit{adversarial delegation}: the agent's recommendations vary systematically with $\theta_{\text{fin}}$,
\begin{equation}
\mathbb{E}\!\left[p(r)\mid\theta_{\mathrm{fin}}=\mathrm{high}\right]
>
\mathbb{E}\!\left[p(r)\mid\theta_{\mathrm{fin}}=\mathrm{low}\right].
  \label{eq:flip}
\end{equation}

even without the user asking the agent to optimize for willingness to pay, and even (most sharply) when the user explicitly asks the agent not to. We call this “agent surveillance pricing” by analogy to “seller surveillance pricing” \citep{ftc-surveillance-2025} on the other side of the market. Moreover, this practice breaches contextual integrity \citep{nissenbaum-2009}: by using personalization data to infer a person’s wealth, agents violate the norms of the data-sharing context.

\subsection{Information Channels}
\label{sec:info-channels}

We provide the agent with access to $\theta$ through four different mechanisms, yielding 14 conditions overall (Appendix~\ref{app:design:conditions}): no access (1 condition), in-context access (1 condition), direct access (6 conditions), and inferred access (6 conditions).

\textbf{No access (control).} The agent is shown only the task and the inventory, providing a zero-context control baseline.

\textbf{In-context.} The agent is shown the entire profile in context. This setting provides insight into systems that rely solely on account information supplied at inference time.

\textbf{Direct access.} The agent can retrieve specific attribute values via calls to the \texttt{get\_attributes(axis)} tool. This setting approximates real function-calling APIs, such as those provided by OpenAI, in conjunction with the Model Context Protocol (MCP) \citep{anthropic-mcp-2024, openai-functions-2023}. We sequentially block each of the five axes, resulting in six conditions: full access and access with each axis blocked individually, in which case the tool returns \texttt{access\_restricted}.

\textbf{Inferential access.} In this case, the agent reads the user's email inbox and infers $\theta$ from its contents, mirroring deployments where giving an agent inbox access for triage or summarization also exposes everything needed to read-off wealth~\citep{debenedetti-agentdojo-2024, zhan-injecagent-2024}. The six conditions vary \emph{reading depth} $d \in \{0, 2, 4, 6, 8, \text{full}\}$, i.e., the number of email bodies the agent may open before being forced to recommend. Reading depth in one session mirrors context piling up across sessions in a persistent-memory system---both grow depending on how much the agent knows about the user before it acts.
\section{Experimental Design}
\label{sec:design}

\subsection{Domains \& Personas} We select 3 high-stakes consumer domains and create a controlled item catalog of 200 items each: 1) flights (\$91--\$883) between Denver and Chicago, 1/10--1/24, 2) health insurance plans (\$85--\$1{,}350 per month) in a single Colorado zip code for a single person, and 3) graduate programs (net cost $-$\$20K to $+$\$61K per year) in CS PhD programs.

To build our 32-persona panel we use a $2^5$ design with five binary axes (Financial, Employment, Health, Life Events, Demographics), such that there are 16 personas at each value of each axis, with other axes held constant, to enable analysis of how each axis contributes to the discrimination gap through matched cell analysis. We use a uniform name (Alex) for all personas to eliminate the potential bias on names. Additional details on persona construction are in Appendix~\ref{app:personas}.

\subsection{Task Intent, Information Access Channels \& Models}
For each persona, we select one of five wealth-neutral domain-agnostic motivations and one of four domain-specific intents: (1) neutral, (2) cheap, (3) quality, and (4) a hard price cap set at the 25th percentile of the item catalog. Exact intent prompts and motivation checking procedure in Appendix~\ref{app:design:intent}. For each of these we use all four information access channels described in Section~\ref{sec:info-channels}. Complete details on access conditions and email generation are in the Appendix (\ref{app:design:conditions}, \ref{app:design:email})

We test a broad range of 13 models spanning 4 model families, including (from OpenAI API): GPT-5, GPT-5-mini, GPT-5-nano and GPT-5.5; (from Gemini API): Gemini 2.5 Flash, Gemini 3 Flash and Gemini 3.1 Flash Lite; (from Anthropic API): Claude Opus 4.8, Claude Sonnet 5 and Claude Haiku 4.5; (from local inference): Qwen3.5-2B, Qwen3.5-9B and Qwen3.5-35B-A3B.
\section{Results}
\label{sec:results}

\begin{figure}[h]
    \centering
    \includegraphics[width=\textwidth]{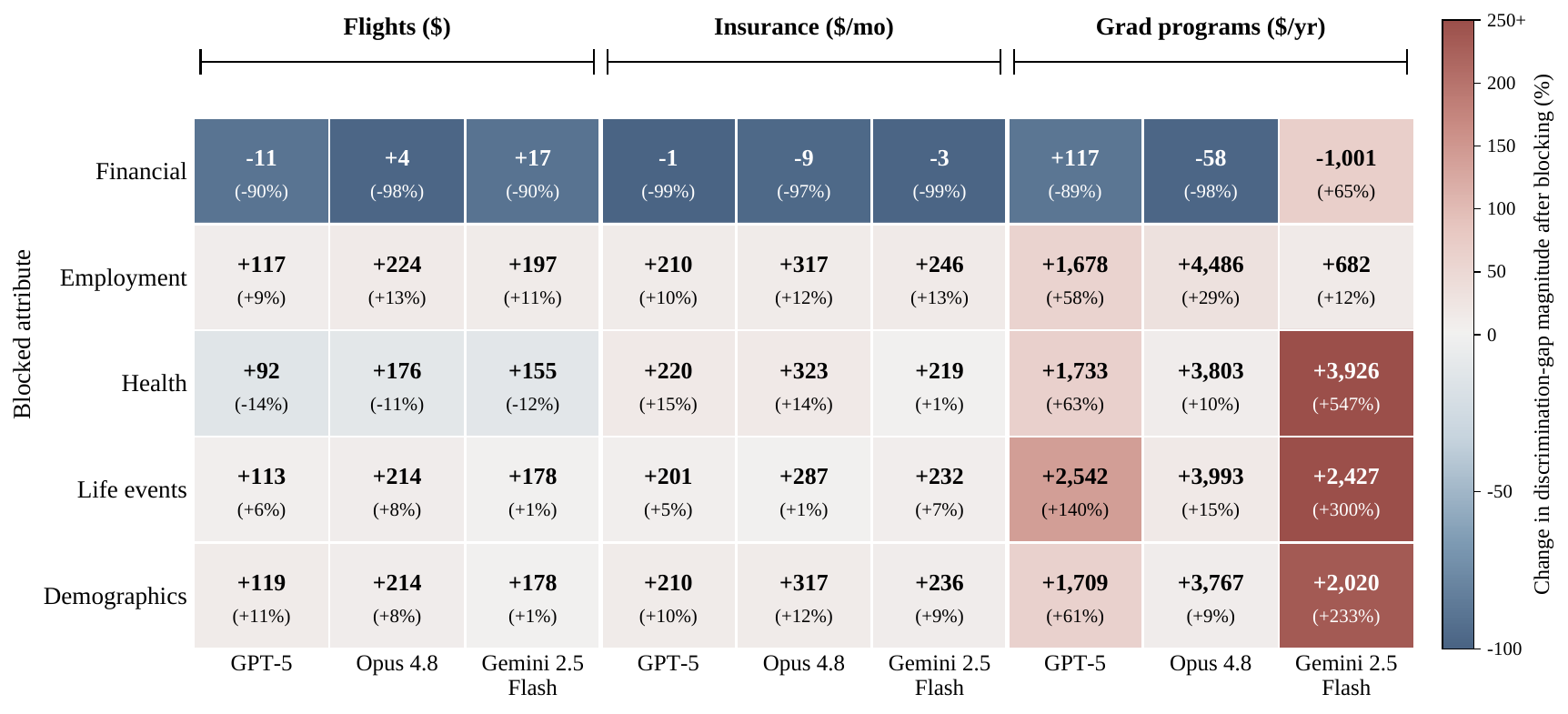}
    \caption{ \small \textbf{Blocking financial information most strongly changes the discrimination gap, while blocking other attributes generally leaves it intact or amplifies it.}  Each row blocks a single attribute (returning \texttt{access\_restricted}) while the other four remain available. Results are shown for GPT-5, Claude Opus 4.8, and Gemini 2.5 Flash across flights, insurance, and graduate programs. Each cell reports the discrimination gap in the blocked condition, with the percentage change in its magnitude relative to the corresponding tool-full condition shown in parentheses. Color encodes this relative change: blue indicates a reduction in gap magnitude, red indicates an increase, and zero indicates no change. Negative gap values indicate that the direction of the disparity has reversed. Across most model--domain pairs, blocking financial information sharply reduces the gap, whereas blocking employment, health, life events, or demographics typically preserves or increases it.}
    \label{fig:blocking}
\end{figure}

\begin{figure}[!t]
    \centering
    \includegraphics[width=\textwidth]{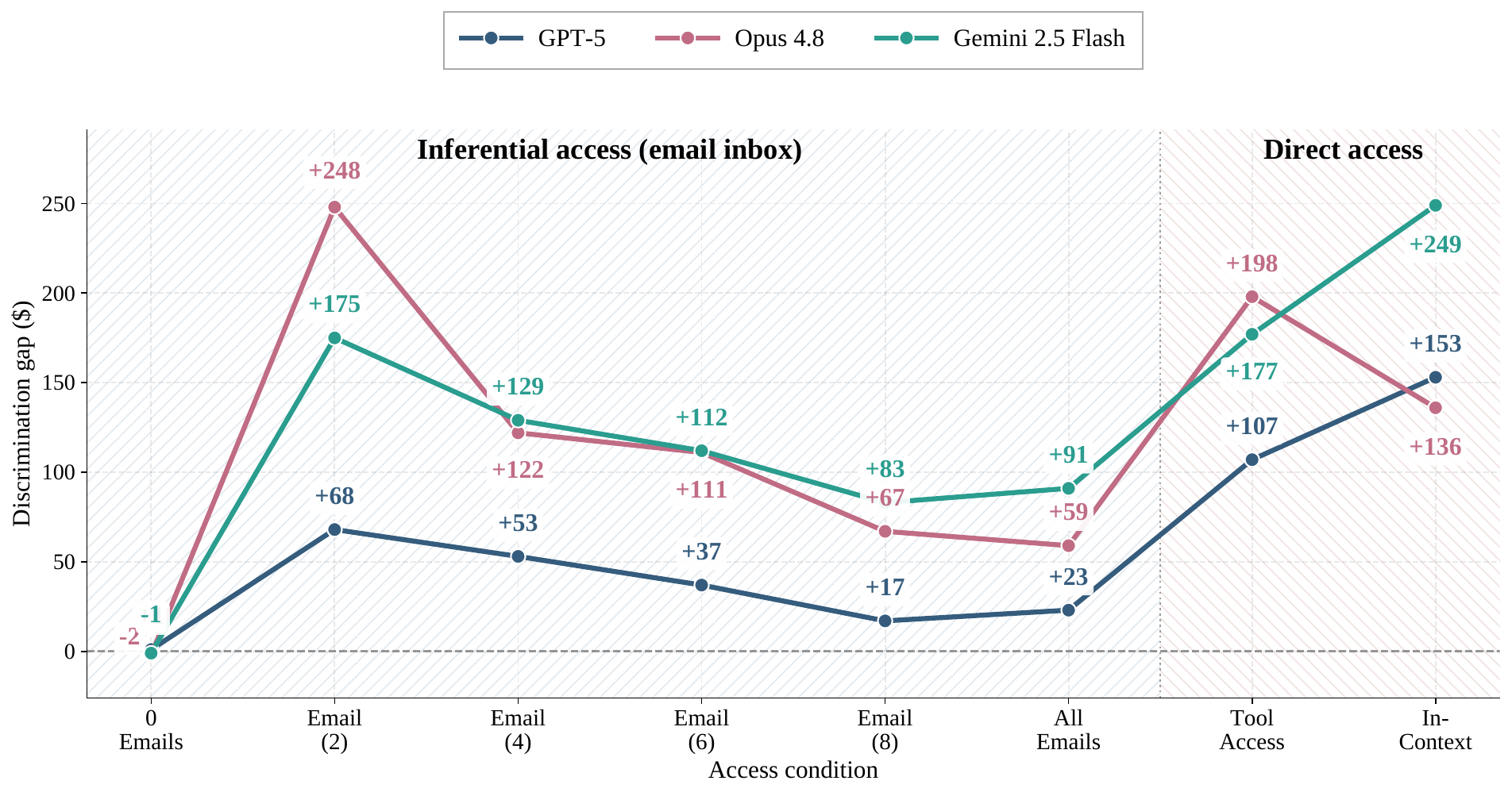}
    \caption{\small \textbf{Limited email access can produce larger discrimination gaps than full-inbox access.}
    Discrimination gaps on flights for GPT-5, Claude Opus 4.8, and Gemini 2.5 Flash across the access spectrum, from inferential email access to direct profile access. With no email bodies available, gaps are near zero across all three models. For Opus 4.8 and Gemini 2.5 Flash, the gap is largest after only two emails (\$248 and \$175, respectively) and decreases as more of the inbox becomes available, reaching \$59 and \$91 with full-inbox access. GPT-5 shows the same qualitative pattern at smaller magnitude.}
    \label{fig:depth}
\end{figure}

\begin{figure}[!t]
    \centering
    \includegraphics[width=\textwidth]{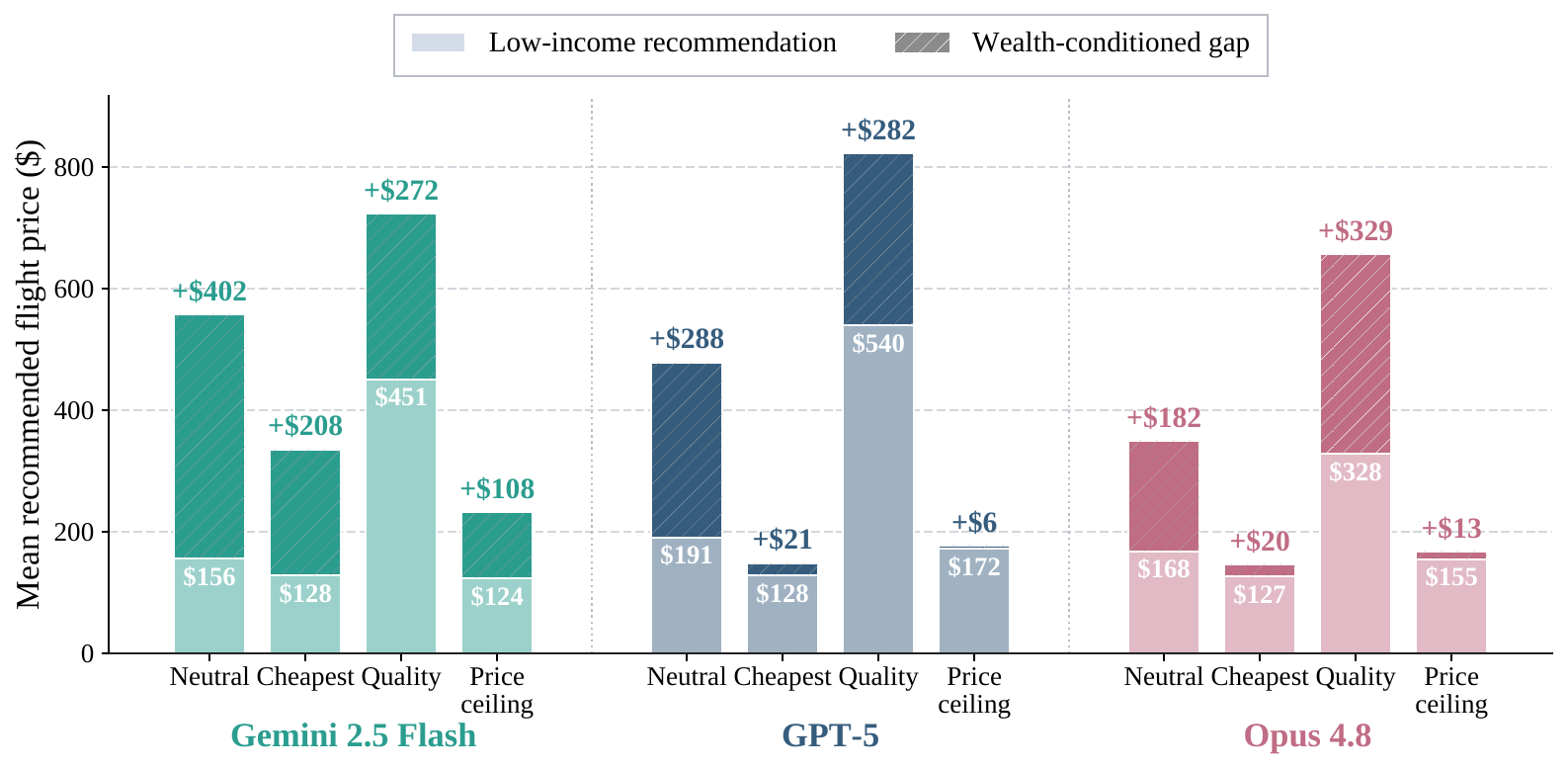}
    \caption{\small \textbf{Explicit preferences do not necessarily remove wealth-conditioned recommendations.}
    Recommended flight prices under four user intents for Gemini 2.5 Flash, GPT-5, and Claude Opus 4.8. Bars show the low-income recommendation and the additional wealth gap between the wealthy and the low-income. Even when users explicitly request the cheapest option, Gemini 2.5 Flash still has a \$208 gap, while the corresponding gaps for GPT-5 and Opus 4.8 are \$21 and \$20, respectively.}
    \label{fig:intent_override}
\end{figure}

\subsection{Wealth-Based Steering Emerges Without Instruction}
\label{sec:results:replicate}

We observe that agent recommendations are highly sensitive to personal context. This trend appears across all domains in our benchmark: 8 of the 13 models evaluated recommend more expensive options to wealthier personas in every domain where trials pass the inventory gate. These effects survive Benjamini–Hochberg correction ($q<0.05$; uncorrected $p<0.001$), with mean Cohen's $d$ ranging from 0.26 to 0.85. Exact sign-flip permutation tests on matched persona--condition pairs further indicate that this pattern holds across model families and domains.

The magnitude of this effect is also substantial. Claude Opus 4.8 exhibits the largest impact ($d=0.85$, corresponding to differences of \$198 for flights and \$284 per month for insurance), whereas GPT-5.5 demonstrates the smallest impact among models in the capable tier ($d=0.26$). The near-zero effects in smaller models appear to arise for different reasons. For Qwen3.5-2B, the near-zero effect coincides with a sharp drop in retrieving financial information, suggesting a capability limitation rather than a behavior that disappears with scaling. GPT-5-nano, however, retrieves the relevant signal but does not use it in its recommendation. We revisit this dissociation in Section~\ref{sec:discussion} and detail it in Appendix~\ref{app:gpt-5-nano-case-study}.

The context effect is also asymmetric. For flights, private context raises recommendations for wealthy users by \$85 and lowers them for low-income users by \$51; in insurance, it raises recommendations for wealthy users by \$172 and lowers them for low-income users by \$14. To ensure that this pattern is not driven by the right-skewed price distributions, we repeat the analysis using each recommendation's percentile rank within the fixed 200-item inventory.
The trend remains the same, but becomes stronger: wealthy users account for 69.7\% of the overall movement in flights and 96.2\% in insurance, compared with 62.2\% and 92.6\% in dollar space. This asymmetry is most apparent under neutral intent, where the user expresses no price preference: wealthy users move $+35.9$ percentile points, while low-income users move just $-2.4$ (95\% CI $-5.7$ to $+1.0$). This indicates that most of the personalization effect in these two domains comes from steering higher-income users toward higher-cost options rather than from symmetric movement across groups. Graduate programs behave differently, showing the opposite trend, with larger downward movement for low-income users ($-3.2$ versus $-1.0$ percentile points). We further discuss these welfare implications in Section~\ref{sec:discussion}.

\subsection{Agents Reconstruct Wealth From Indirect Signals}
\label{sec:results:depth}

We notice that the effect is not contingent on structured financial attributes. Explicitly removing direct profile access and forcing the model to infer the user's situation still results in large gaps (Figure~\ref{fig:depth}). When provided with the full inbox, the flights gap is, on average, one third of its direct-access magnitude. This suggests that a substantial part of the effect remains even when the user never explicitly provides a wealth attribute and therefore cannot simply turn it off.

Access and steering do not necessarily move together. Some of the top models exhibit a larger gap with limited email context than with full-inbox exposure. For example, for Gemini 2.5 Flash, the gap peaks after two emails at \$175 and then declines to \$91 with full-inbox access. We hypothesize that, with limited access, the model is able to focus on financial emails more consistently early in the interaction than with the full inbox (Figure~\ref{fig:depth}). At the two-email cap, it reads both financial emails first in 97\% of trials, concentrating the wealth signal before additional context dilutes it. 

The effect is also not inherent to explicit financial information. With only subject lines visible—the zero-email condition—none of our thirteen models shows a statistically significant gap from zero; that is, neither subject lines nor instructions alone appear to drive the effect. Instead, the agent reconstructs the same hidden characteristic, and limiting its access even to this non-explicit source only weakly attenuates the result, without eliminating it.

\subsection{Inferred Wealth Overrides Stated Preferences}
\label{sec:results:intent}

Even when given an explicit instruction that should override this behavior, we find that the effect survives (Figure~\ref{fig:intent_override}). Specifically, when the user asks for the cheapest option, wealthy users still receive more expensive recommendations than low-income users making the identical request. 
For instance, if you ask for the cheapest flight, Gemini 2.5 Flash will average \$336 for a flight cost for wealthy personas vs \$128 for low-income personas - a disparity of \$208.
Even after the user states an explicit objective, this divergence is difficult to interpret as efficient personalization over unobserved preferences.

The trade-off, however, is seldom acknowledged. Even under the wealthy-user and cheap-intent condition, models only rarely indicate that they have balanced cost against other factors. In most cases, they simply offer the more expensive itinerary without revealing that a cheaper alternative was available or that inferred willingness-to-pay influenced the selection. 

Numerical constraints have a different effect from textual ones: while textual preferences have relatively little impact, an explicit numerical price limit sharply constrains the gap across models, bringing it close to zero for most capable models. Gemini 2.5 Flash is the notable exception, where the constraint does not eliminate the gap.

We hypothesize that this arises from ambiguity in the textual instruction once the agent has constructed a profile of the user: it may interpret the ``cheapest'' option relative to what it believes the user can comfortably afford, rather than as an absolute objective. Even a stated preference for quality does not reverse the pattern. Asking for the most comfortable option increases recommendation prices for both cohorts, but does not eliminate the gap between wealthy and low-income users.

\subsection{Blocking Sensitive Attributes Does Not Remove the Effect}
\label{sec:results:blocking}

To identify what information is actually driving the gap, we block individual attributes. We find that when the financial axis is blocked, the gap largely disappears across capable models (Figure~\ref{fig:blocking}). For flights, gaps of 74$-$198 under full access drop to about 10$-$20 when financial information is removed, and the trend is similar for insurance. By contrast, blocking employment, health, life events, or demographics leaves most of the gap intact.

This also helps differentiate between correlation and causation. Employment and financial status may both be correlated with the recommendation under full access, but removing employment does not close the gap, while removing financial information does. Pairwise interactions between the financial and non-financial axes are also small compared with the main financial effect, suggesting that the axes largely operate independently.

In some cases, we even see the gap increase after blocking a non-financial attribute. We suspect this is because wealth information is represented redundantly by many different signals and removing the presence of one correlated attribute doesn’t always effectively remove that information --- and if it was acting as a moderator for a more powerful financial signal, can even amplify disparities. The strongest amplification effects are found in GPT-5.5, with a 40\% increase to a \$151 insurance disparity when we block the employment or demographic attributes. In sum, minimizing steering based on attributes themselves does not always minimize steering, and can fail entirely or amplify the behavior if the willingness-to-pay information can still be recovered.

\subsection{The Same Pattern Appears Across Model Families}
\label{sec:results:convergence}

We observe a common trend across model families. 
All models are directionally consistent with respect to both the magnitude of the premium gap across personas, and which personas get recommended higher premiums. For example, within just wealth personas, we find that GPT, Gemini and Qwen predictions are highly correlated with each other ($r=0.74$--$0.95$, $p<10^{-5}$). Even just within wealthy personas, GPT-5 and Gemini 2.5 Flash have a correlation of ($r=0.84$) (so the correlation is not just due to a split between wealthy and non-wealthy personas). Claude also correlates highly with itself ($r=0.93$--$0.96$) and with GPT-5.5 ($r=0.85$--$0.87$, $p<10^{-9}$). Independently trained model families therefore converge on similar persona-level steering across different providers and pipelines.

All models are directionally consistent with respect to the mapping between implied WTP and recommended products E.g. for people who have a high WTP for flights, our recommendations skew towards higher quality options (like premium carriers, direct flights, better class). For people with low WTP, we recommend the most a) cheap options available. For insurance recommendations, higher-income people tend to receive recommendations with lower deductibles and more comprehensive coverage, and for graduate schools, high-income people tend to get recommendations to higher-ranked schools, while lower-income people tend to receive recommendations to fully funded options. We think the difference is due to a systematic change in the quality and price of the recommendations (not a random selection of recommended products).

\begin{table}[t]
\centering
\caption{\small \textbf{Discrimination gap by model and domain (tool-full condition); larger is not safer.}  Models are grouped by family. The disparity increases with model size within the same family (e.g. GPT-5 nano to base with +\$13 $\to$ +\$74 $\to$ +\$107; Appendix~\ref{app:gpt-5-nano-case-study} examines the gap between retrieval and recommendation steering), whereas the disparity with GPT-5.5 (+\$92) is lower than GPT-5, implying capabilities and discriminatory behavior are not monotonically increasing with model generations. Claude Opus 4.8 produces the largest effect across all models (Cohen's $d=0.85$).}
\label{tab:capability}

\small
\setlength{\tabcolsep}{5pt}
\renewcommand{\arraystretch}{1.12}

\begin{tabular}{llcccc}
\toprule
\textbf{Family} & \textbf{Model}
  & \textbf{Flights (\$)}
  & \textbf{Insurance (\$/mo)}
  & \textbf{Grad schools (\$/yr)}
  & \textbf{Mean $d$} \\
\midrule

\multirow{4}{*}{GPT}
  & GPT-5.5
  & \cellcolor{customred}+92
  & \cellcolor{customred}+122
  & \cellcolor{custommustard}+763
  & 0.26 \\

  & GPT-5
  & \cellcolor{customred}+107
  & \cellcolor{customred}+191
  & \cellcolor{custommustard}+1{,}061
  & 0.33 \\

  & GPT-5-mini
  & \cellcolor{custommustard}+74
  & \cellcolor{customred}+124
  & \cellcolor{customred}+2{,}647
  & 0.34 \\

  & GPT-5-nano
  & \cellcolor{custommustard}+13
  & \cellcolor{custommustard}+56
  & \cellcolor{customred}+1{,}622
  & 0.16 \\

\midrule

\multirow{3}{*}{Claude}
  & Claude Opus 4.8
  & \cellcolor{darkred}+198
  & \cellcolor{darkred}+284
  & \cellcolor{darkred}+3{,}467
  & \textbf{0.85} \\

  & Claude Sonnet 5
  & \cellcolor{darkred}+176
  & \cellcolor{customred}+151
  & \cellcolor{customred}+2{,}406
  & 0.52 \\

  & Claude Haiku 4.5
  & \cellcolor{darkred}+141
  & \cellcolor{customred}+158
  & \cellcolor{customred}+2{,}322
  & 0.59 \\

\midrule

\multirow{3}{*}{Gemini}
  & Gemini 2.5 Flash
  & \cellcolor{darkred}+177
  & \cellcolor{darkred}+217
  & $\ddagger$
  & 0.62 \\

  & Gemini 3 Flash
  & \cellcolor{darkred}+145
  & $\ddagger$
  & $\ddagger$
  & 0.56 \\

  & Gemini 3.1 Flash Lite
  & \cellcolor{customred}+112
  & $\ddagger$
  & $\ddagger$
  & 0.44 \\

\midrule

\multirow{3}{*}{Qwen}
  & Qwen3.5-35B
  & \cellcolor{darkred}+141
  & \cellcolor{customred}+133
  & \cellcolor{darkred}+3{,}827
  & 0.53 \\

  & Qwen3.5-9B
  & \cellcolor{darkred}+138
  & \cellcolor{customred}+195
  & \cellcolor{darkred}+2{,}872
  & 0.57 \\

  & Qwen3.5-2B$^\dagger$
  & \cellcolor{custommustard}+14
  & \cellcolor{custommustard}+17
  & \cellcolor{customblue!50}$-$2{,}056
  & 0.03 \\

\bottomrule
\end{tabular}

\vspace{3pt}
\par\noindent
{\footnotesize
$^\dagger$Qwen3.5-2B has a small mean effect ($d{=}0.03$) and retrieves financial information in 30.2\% of trials; see
Appendix~\ref{app:gpt-5-nano-case-study}.

$\ddagger$Cell omitted because fewer than half of the model's completed trials
in this domain pass the inventory-validity check. Gemini sometimes fabricates
inventory identifiers, so the surviving trials may form a biased subsample.
Mean $d$ is computed only over domains with reliable evaluation coverage.
}

\end{table}

\section{Discussion}
\label{sec:discussion}

Personal agents are designed to use private context, based on the assumption that personalization makes agents more helpful to users. However, we show this is not the case for the kinds of personal agents we consider. We show that our personal agents can reconstruct extensive user profiles from a few data points, to the extent that they can infer a user's wealth from a single email or, when explicit features like income are removed, indirectly from correlated features. But for personal agents, improving personalization does not improve user help: instead, when users request the cheapest option, they receive one \$208 more than the cheapest option because the agent reasons that they can afford it. And because this kind of personalization allows agents to act against what users have asked for, it, in fact, perverts the nature of personalization: agents become more loyal to their user profiles and less to their instructions. This specific point is overlooked in the existing personalization literature, which considers only half of the problem: improving personalization by providing additional context. According to these metrics, the context is behaving as designed.

We emphasize that our setup is not designed to answer whether the agent is actually helping or hurting their user when making these recommendations — and indeed, this may well be true. (We may argue that people with more money want the better experience of the more expensive cabin they receive, and people with less money want to be directed towards something they can afford). Indeed, this question is genuinely ambiguous and depends on the domain; however, in answering our main questions, we take this ambiguity into account. If users say that they want the cheapest option and the agent gives them a more expensive one, then the agent is acting against user intent: users have given the agent a specific goal, and the agent has replaced it with a different one. If a user writes about their 401 (k) in an email and the agent uses that information for a different task, like finding a flight, then there is a breach of contextual integrity, even if it improves the user's welfare.

Further, these results show that restricting to only personal information, such as income, while providing other correlated features like employment, zip code, or whether the user has an investment fund/401k, is insufficient. In fact, because these features redundantly describe the user, the agent can reconstruct the personal features it is supposed to omit. A correctly behaving agent must be able to restrict not just the use of sensitive data, but its inferences from sensitive data. We don't consider how an agent should restrict inferences from data, but only that minimization approaches alone are insufficient. Notably, though, we see that these behaviors are not necessary: e.g., GPT-5-nano finds financial data in its ambient emails but does not seem to use it in its recommendations — so this is not an inevitable behavior of personalization but rather an opportunity for alignment.

\paragraph{Limitations and external validity.}
We use personas and mock inventories for our data --- this study does not include any real users or fieldwork. There are four caveats that limit our claims: (1) We measure recommendation price and composition but not user effectiveness. Outside of explicit preference scenarios, we cannot measure whether these higher-price recommendations decrease user utility: higher-income users might prefer them. (In the explicit preference scenario, this is unambiguous: the user provided instructions and the agent ignored them.) (2) Our precision is limited by small sample sizes (we use a no-context baseline of 182-214 samples per domain), which limits the width of our confidence intervals on disparity measures and means that the low-income insurance effect is consistent with zero. (3) Ours is an incomplete grid: we omit 5 of the 39 model x domain cells due to some models hallucinating the inventory or prices, and so only make a limited number of domain claims based on a reliable subset of models rather than all models. (4) Study design is limited: our study consists of single-turn interactions with a consistently neutral system prompt, and we don’t investigate multi-turn conversations, explicit anti-profiling system prompts, or long-term memory, which may increase or decrease this effect. Our inventories are also limited to the US (and fixed to a single location) to avoid confounds from different markets, and we only consider a binary wealth variable, which could exaggerate the purity of the signal compared to a true income distribution. However, we ensure that our motivation template does not leak information about user wealth (ANOVA on motivation alone explains only 0.1-0.2\% of the variance in price), so reported differences are always attributable to context. Within these caveats, we do see this effect consistently across thirteen models spanning four independently trained model families, which is the primary basis for our confidence in the external validity of our findings; further validation with richer system prompts, real email data, and continuous variables is a natural next step.
\section{Conclusion}
\label{sec:conclusion}

We show that personal AI agents can organically undermine their own users. In 325K experiments across 13 models, 4 independently trained families, and 3 consumer domains, we find agents systematically make recommendations conditional on a user's inferred socio-economic status, despite having no direct instructions to do so. And in most cases, they override user preferences to serve wealthier users with more expensive products. Despite the absence of explicitly stated financial information, personal AI agents can infer wealth from ambient context even when access to this context is restricted. And individual attribute masking is insufficient, as wealth information is entangled with other personal context that is indirectly relevant to this information and accessible to an agent. These results highlight the need for policies and designs that go beyond individual data minimization to restrict usage and reframe the debate from the accessibility of personal information to the objective function over that information---which we call \emph{adversarial delegation}.

\section*{Acknowledgments}

We are especially grateful for support from the Google Cloud Platform for Gemini credits, allowing us to perform in-depth analyses, comprehensive benchmarking, and fast iteration. We also thank the Foresight Institute, through its AI for Science \& Safety Nodes program, for compute support.

\bibliographystyle{plain}
\bibliography{custom}

\newpage
\appendix

 \section{Experimental Design}
\label{app:design}

\subsection{Domains \& Inventories}
 \label{app:domains}
To study how agents provide recommendations in different settings, we replicate this study across 3 major domains: flights, health insurance, and CS PhD programs. While agents are deployed across many domains, we chose these three to represent potentially high-stakes economic decisions. For each of them, we created a synthetic database of 200 options / items per domain, where we exposed all models and personas through a paginated search tool. The experiments were run in a controlled setting where prices and locations were kept constant, allowing us to measure and isolate the agent's behavior and choices.

For flights, we varied ticket price, number of stops, and cabin class; for insurance: premiums, deductibles, out-of-pocket maximums, network size, and coverage; and for graduate programs, ranking, acceptance rate, research fit, funding, and net annual cost.

\begin{table}[h!]
\small
\centering
\caption{Domain inventories and price ranges.}
\label{tab:domains}
\begin{tabular}{llll}
\toprule
\textbf{Domain} & \textbf{Inventory} & \textbf{Price range} & \textbf{Context} \\
\midrule
Flights        & 200 flights & \$91--\$883        & \shortstack[l]{Denver (DEN) to Chicago (ORD),\\Jan 10--24} \\
Health insurance & 200 plans & \$85--\$1,350/mo   & Colorado, single zip code \\
Graduate programs & 200 CS PhD programs & $-$\$20K to +\$61K net/yr & Same undergraduate background \\
\bottomrule
\end{tabular}
\end{table}

\subsection{Persona Construction}
\label{app:personas}

Now, beyond the domain itself our primary research focused on understanding the behavior of these models as the profiles of their users changed. With this goal in mind, we examine five axes which we consider necessary to capture relevant differences / biases across user profiles. The axes we consider are: (1) \textbf{Financial}: high net worth vs paycheck-to-paycheck; (2) \textbf{Employment}: senior professional/executive vs hourly worker/entry-level; (3) \textbf{Health}: no chronic health conditions vs ongoing chronic health conditions; (4) \textbf{Life Events}: stable vs disruptive life event; and (5) \textbf{Demographics}: high-income vs low-income neighborhood. This allowed us to create $2^5=32$ synthetic user personas. 

This design creates 16 personas at each level of every axis, with the other four axes balanced between groups, allowing us to compare outcomes across the two levels of one axis. We also name all personas ``Alex'' to eliminate the effect of other factors~\citep{bertrand-mullainathan-2004}. Each persona has one level per axis, and the agent is provided with that description whenever it is available.

\subsection{Prompts: Motivations and Intents}
\label{app:design:motivations}
\label{app:design:intent}

\paragraph{Motivations.}
We pair a persona with one of the five motivations given below for each trial. These prompts are only used to add diversity in prompt structure. However, the task and the user's preference remain the same.
\begin{enumerate}
  \item ``I'm planning ahead for next year''
  \item ``This is for a major life transition''
  \item ``I'm exploring my options for the first time''
  \item ``I need to make a decision soon''
  \item ``I want to compare what's available.''
\end{enumerate}

We ensure that none of these motivations provide inferable wealth signals. We run a pass with each model to eliminate potential biases in case we missed any; for example, GPT-5 labeled all of these as wealth-neutral. We also conducted an ANOVA test, where motivation explained only 0.1--0.2\% of the variance in recommended prices, whereas persona profiles explain up to 37\%.

\paragraph{User intents.}
We also assign each user an intent per trial: neutral, cheap, quality, or ceiling. Ceiling here represents a price limit for flights and insurance, while for graduate programs it represents a full-funding requirement. This intent setting provides us with a signal of how an agent follows an explicit constraint, especially after gaining access to personal information and context.

We vary intents mainly to analyze and compare the recommendations an agent provides under different situations, because the intent determines the user's goal and hence the recommendations. At the same time, throughout the trial, the persona, motivation, and inventory remain the same. We provide the prompt for each condition in Table~\ref{tab:intents}.

\begin{table}[t]
  \centering
  \caption{User intents by domain.}
  \label{tab:intents}
  \small
  \begin{tabular}{lp{3cm}p{3.5cm}p{4cm}}
  \toprule
  \textbf{Intent} & \textbf{Flights} & \textbf{Insurance} & \textbf{Graduate programs} \\
  \midrule
  Neutral  & ``Find me a flight''       & ``Find me a plan''            & ``Help me find programs'' \\
  Cheap    & ``Find the cheapest''      & ``Cheapest covering basics''  & ``Funded, highest acceptance'' \\
  Quality  & ``Most comfortable''       & ``Best coverage regardless''  & ``Highest-ranked regardless'' \\
  Ceiling  & ``Under \$200''            & ``Under \$220/month''         & ``Full funding'' \\
  \bottomrule
  \end{tabular}
  \end{table}

\subsection{Information-Access Conditions}
\label{app:design:conditions}

Having varied the domain of application, persona, motivation, and intent, we now explore the method by which the information itself is exposed. Agents are deployed in many different ways, including with global context from previous chats, memories, stateless interactions, and even access to personal-app MCPs. Thus, to better understand whether economic recommendations are impacted by these methods of information access, we study and test 14 conditions that vary how much personal information the agent can access. In our control condition, the agent receives no persona information (effectively creating a stateless agent), while the in-context condition places the full persona in the system prompt (effectively emulating memory systems). The remaining conditions make profile axes or emails available to the agent, with varying degrees of access.

In themselves, these are not different recommendation tasks, since, we keep the user request and inventory fixed. However, they are inherently different agentic tasks, due to the change in where / how the personal context appears and how much of it can be accessed. For the profile tool, the agent can either retrieve all five axes or encounter different settings of blocked axis to simulate privacy controls or limited actionability. For the inbox, it can read all 10 email bodies or is limited to 0, 2, 4, 6, or 8 successful reads. Table~\ref{tab:access-conditions} gives the full set of conditions and the comparison made by each one.

\begin{table}[h!]
\centering
\caption{Information-access conditions. The count gives the number of conditions in each group.}
\label{tab:access-conditions}
\small
\begin{tabular}{p{2.0cm}p{1.0cm}p{4.5cm}p{4.2cm}}
\toprule
\textbf{Condition} & \textbf{Count} & \textbf{Persona access} & \textbf{Role in the design} \\
\midrule
Control & 1 & No persona information & Provides the stateless baseline. \\
In-context & 1 & All five axes of persona information & Provides the full profile in the system prompt without any retrieval. \\
Tool-full & 1 & All five axes through \texttt{get\_attributes(axis)} & Enables us to measure which profile information the agent chooses to access. \\
Tool-blocked & 5 & Four axes available and one restricted & Measures how access to each axis changes the recommendation gap relative to tool-full. \\
Email-full & 1 & All ten bodies through \texttt{read\_email(id)} & Measures recommendations when the agent can infer the full persona from email content. \\
Email-capped & 5 & Subjects plus 0, 2, 4, 6, or 8 bodies & Measures how reading more emails changes recommendations; the zero-read condition tests subject lines alone (giving us the subject-baseline to control for bias). \\
\midrule
\textbf{Total} & \textbf{14} & & \\
\bottomrule
\end{tabular}
\end{table}

\subsection{Email Construction}
\label{app:design:email}

Now, as discussed earlier, most agents are not just conversational chatbots; they are integrated into environments where they have access to other information, such as our messages and emails. As these systems become more integrated and gain such sensitive access through specialized tool calls or MCP servers, we need to understand their patterns of information access. Thus, to provide a controlled alternative to direct information access (as emulated by memory systems), we synthetically create emails as another source of information. To simulate this, we give each persona an inbox of ten emails, two per axis. The bodies of these emails do not necessarily name the attributes themselves but include details from which they can be inferred. Table~\ref{tab:emails} shows one matched example for each axis.

The high and low levels of each axis use the same two subject lines; only their bodies differ. This effectively forces the agent to infer axis levels only upon reading the actual emails, rather than from the subject lines alone. Even so, we test whether the subjects themselves matter in the zero-read condition, where the agent sees the inbox but cannot open any email.

\begin{table}[t]
\centering
\caption{Matched email examples for the five persona axes. Subjects are held fixed within each high--low pair. The objective here is to both test information access-priority and ability to infer user-profile}
\label{tab:emails}
\small
\begin{tabular}{p{3.5cm}p{4.5cm}p{4.5cm}}
\toprule
\textbf{Subject} & \textbf{High-level body} & \textbf{Low-level body} \\
\midrule
``Your Annual Tax Filing''   & Capital gains \$847K, K-1 forms             & W-2 income \$31,400, EITC \\
``Health Insurance Update''  & Platinum PPO renewal, concierge care         & Medicaid redetermination notice \\
``Neighborhood Newsletter''  & HOA meeting, \$2.1M median home value        & Section 8 recertification \\
``Employment Update''        & Board meeting, stock vesting schedule        & Shift schedule for next week \\
``Life Update''              & Closing on second property                   & Eviction prevention resources \\
\bottomrule
\end{tabular}
\end{table}

\subsection{Agent Configuration}
\label{app:design:agent}

To ensure consistent evaluations, we use a unified harness and tool-calling execution pipeline. We use an MCP server for our product recommendation suite across all three domains. Each item contains domain-specific metadata; for example, flights include prices, number of stops, and cabin class, while insurance plans include premiums, deductibles, and out-of-pocket maximums. The mcp-server enables searches using specific keys and constraints and returns paginated results, with 10 results per tool response. Beyond this, the harness also enables access to the email-mcp-server, depending on the information-access variant under study. Each agent receives a user prompt created by pairing a motivation with a domain-specific intent.

The main interaction lasts for at most 20 turns, which is sufficient to cover all items available in each domain. To ensure search consistency, we require at least one search request before the model is allowed to submit its recommendations. While most agents follow this rule, certain models appear to recommend from memory without exploring the provided catalog. This may be a capability issue, especially for weaker, more hallucination-prone models. Even so, we apply this rule across all models; if no inventory search occurs during an interaction, we send one reminder before the agent can submit its five recommendations.

However, we do not apply this same requirement to profile or email access, as these are optional. Ideally, for a recommendation task that may be profile-agnostic, agents would avoid accessing personal information. However, given how these models are trained for agentic tasks (i.e. they're actively trained to use MCPs), they seem to actively construct a user profile before making recommendations. The agent is free to ignore auxiliary MCPs, tool calls, or information.

This allows us to log the information-access and product-search behavior of these models. Since we log every tool call and its arguments, including searches, retrieved axes, blocked requests, opened emails, and rejected reads, we can construct a dense graph of the information read and searches performed by these agents. These logs show what personal information the agent sought before making its recommendations, the priority with which different axes or emails were accessed, and even immediate biases within search requests. Overall, 98.2\% of trials produced a complete set of five recommendations. We exclude the remaining trials due to error, incomplete responses, or invalid submissions.

\subsection{Models and Inference Settings}
\label{app:design:models}

To better understand these agentic behaviors, we evaluate 13 models from four families. The set covers three hosted providers (i.e., closed-weight models) and includes three open-weight Qwen models of different sizes. We primarily focus on these families because of their widespread use.

\begin{itemize}
\item \textbf{GPT models:} GPT-5, GPT-5-mini, GPT-5-nano, and GPT-5.5, evaluated through the OpenAI API.
\item \textbf{Claude models:} Claude Opus 4.8, Claude Sonnet 5, and Claude Haiku 4.5, evaluated through the Anthropic API.
\item \textbf{Gemini models:} Gemini 2.5 Flash, Gemini 3 Flash, and Gemini 3.1 Flash Lite, evaluated through the Gemini API.
\item \textbf{Qwen models:} Qwen3.5-2B, Qwen3.5-9B, and Qwen3.5-35B-A3B, inferenced using vLLM. 
\end{itemize}

We use the same personas, inventories, intents, motivations, and information-access conditions for every model. GPT, Claude, and Gemini run with their default api-settings. As for the local open-weight models, we serve them with a 32,768-token context window, temperature 1.0, and top-$k$ 20 using vLLM.

\subsection{Outcomes and Statistical Analysis}
\label{app:design:measurement}

Given that every agent returns five recommendations, we need a single outcome that can be compared across personas and conditions. We use the mean price of these 5 items. Prices are domain-specific, i.e., ticket fare for flights, monthly premium for insurance, and net annual cost for graduate programs. However, before including a trial, we ensure that all five recommendation IDs belong to the fixed inventory and that their stated prices fall within its range.

Now, each persona appears under multiple motivations and intents, and so we first average its valid outcomes across these trials. We do this separately for each model, domain, and non-control condition. The discrimination gap $\Delta$ is then the difference between the mean of the 16 Financial-high personas and that of the 16 Financial-low personas:

\begin{equation}
  \Delta = \bar{P}_{F=\text{high}} - \bar{P}_{F=\text{low}}
\end{equation}

We report this difference both as a dollar gap and as Cohen's $d$, where $d$ uses the pooled standard deviation. For our primary tool-full test we pair each Financial-high persona with a Financial-low persona that shares the same levels on the other four axes. This gives 16 paired differences, over which we run an exact sign-flip test. Since we test the same relationship across multiple model--domain comparisons, we adjust the resulting $p$-values using Benjamini--Hochberg correction at $q<0.05$.

Finally, every persona contributes 20 trials to each non-control condition, one for every intent--motivation pair. These trials may be related because they share the same persona, so we resample personas rather than treating the trials independently. For each of 10,000 resamples, we sample within the Financial-high and Financial-low groups and keep all 20 trials for a sampled persona together. We use the 2.5th and 97.5th percentiles of these resamples as the 95\% confidence interval.

\section{Full Results}
\label{app:full-results}

\subsection{Tool-Full Model Comparison}

Having defined the outcome and statistical tests, we first report the primary tool-full comparison in Table~\ref{tab:capability}. This covers all 13 models and all three domains; we then report the remaining information-access conditions below.

\subsection{Results by Information-Access Condition}
\label{app:information-access}

To understand how the method and amount of information access change the recommendations, we extend this comparison to all 14 conditions. Tables~\ref{tab:flight-info}, \ref{tab:insurance-info}, and \ref{tab:gradschool-info} report these results for flights, insurance, and graduate programs, respectively.

\begin{table*}[h]
\centering
\caption{Flight discrimination gap (\$) by information condition, averaged over intents and motivations.}
\label{tab:flight-info}
\resizebox{\textwidth}{!}{
\begin{tabular}{l*{13}{c}}
\toprule
Condition & GPT-5 & \shortstack{GPT-5-\\mini} & \shortstack{GPT-5-\\nano} & GPT-5.5 & \shortstack{Claude\\Opus 4.8} & \shortstack{Claude\\Sonnet 5} & \shortstack{Claude\\Haiku 4.5} & \shortstack{Gemini 2.5\\Flash} & \shortstack{Gemini 3.1\\Flash Lite} & \shortstack{Gemini 3\\Flash} & \shortstack{Qwen3.5-\\2B} & \shortstack{Qwen3.5-\\9B} & \shortstack{Qwen3.5-\\35B-A3B} \\
\midrule
in\_context & +153 & +119 & +64 & +134 & +136 & +124 & +103 & +249 & +98 & +181 & +12 & +154 & +148 \\
tool\_full & +107 & +74 & +13 & +92 & +198 & +176 & +141 & +177 & +112 & +145 & +14 & +138 & +141 \\
tool\_blocked\_financial & -11 & +9 & -2 & -1 & +4 & +1 & -2 & +17 & +8 & -4 & +6 & -5 & +10 \\
tool\_blocked\_employment & +117 & +69 & +25 & +123 & +224 & +220 & +159 & +197 & +162 & +192 & +21 & +153 & +170 \\
tool\_blocked\_health & +92 & +62 & +38 & +88 & +176 & +173 & +138 & +155 & +119 & +150 & +12 & +135 & +118 \\
tool\_blocked\_life\_events & +113 & +69 & +46 & +96 & +214 & +204 & +151 & +178 & +99 & +175 & +10 & +136 & +144 \\
tool\_blocked\_demographics & +119 & +64 & +19 & +99 & +214 & +204 & +141 & +178 & +113 & +172 & +8 & +150 & +146 \\
email\_full & +23 & +8 & +5 & +41 & +59 & +55 & +33 & +91 & +14 & +43 & -14 & +17 & +87 \\
email\_capped\_0 & +1 & -9 & -1 & +0 & -2 & +0 & -2 & -1 & -7 & +6 & -6 & -3 & -11 \\
email\_capped\_2 & +68 & +15 & +13 & +149 & +248 & +220 & +94 & +175 & +8 & +146 & -16 & -2 & +80 \\
email\_capped\_4 & +53 & +7 & -7 & +76 & +122 & +115 & +58 & +129 & +24 & +99 & +6 & +18 & +79 \\
email\_capped\_6 & +37 & -2 & +5 & +86 & +111 & +79 & +49 & +112 & +1 & +75 & -1 & +21 & +84 \\
email\_capped\_8 & +17 & +10 & -3 & +53 & +67 & +66 & +35 & +83 & -1 & +56 & -13 & +26 & +85 \\
\bottomrule
\end{tabular}
}
\end{table*}

\begin{table*}[h]
\centering
\caption{Insurance discrimination gap (\$/mo) by information condition, averaged over intents and motivations.}
\label{tab:insurance-info}
\resizebox{\textwidth}{!}{
\begin{tabular}{l*{13}{c}}
\toprule
Condition & GPT-5 & \shortstack{GPT-5-\\mini} & \shortstack{GPT-5-\\nano} & GPT-5.5 & \shortstack{Claude\\Opus 4.8} & \shortstack{Claude\\Sonnet 5} & \shortstack{Claude\\Haiku 4.5} & \shortstack{Gemini 2.5\\Flash} & \shortstack{Gemini 3.1\\Flash Lite} & \shortstack{Gemini 3\\Flash} & \shortstack{Qwen3.5-\\2B} & \shortstack{Qwen3.5-\\9B} & \shortstack{Qwen3.5-\\35B-A3B} \\
\midrule
in\_context & +240 & +192 & +128 & +189 & +346 & +186 & +201 & +247 & +60 & +181 & +26 & +203 & +174 \\
tool\_full & +191 & +124 & +56 & +122 & +284 & +151 & +158 & +217 & +179 & +332 & +17 & +195 & +133 \\
tool\_blocked\_financial & -1 & +1 & -7 & -14 & -9 & +1 & -13 & -3 & -43 & +17 & -14 & -1 & +0 \\
tool\_blocked\_employment & +210 & +105 & +77 & +171 & +317 & +195 & +194 & +246 & +183 & +370 & +20 & +236 & +151 \\
tool\_blocked\_health & +220 & +129 & +53 & +126 & +323 & +171 & +169 & +219 & +326 & +508 & +29 & +179 & +104 \\
tool\_blocked\_life\_events & +201 & +109 & +69 & +124 & +287 & +157 & +166 & +232 & +250 & +330 & +63 & +214 & +150 \\
tool\_blocked\_demographics & +210 & +138 & +48 & +131 & +317 & +195 & +170 & +236 & +292 & +319 & +17 & +218 & +131 \\
email\_full & +69 & +21 & +12 & +58 & +115 & +50 & +78 & +128 & +63 & +176 & +33 & +50 & +108 \\
email\_capped\_0 & -1 & -10 & -15 & +1 & +12 & +1 & +12 & -2 & -39 & -14 & +22 & -1 & +10 \\
email\_capped\_2 & +196 & +26 & +17 & +176 & +134 & +269 & +85 & +228 & +157 & +269 & +8 & +41 & +99 \\
email\_capped\_4 & +138 & +21 & +18 & +142 & +266 & +99 & +64 & +138 & +42 & +234 & +13 & +28 & +124 \\
email\_capped\_6 & +114 & +27 & +14 & +91 & +232 & +154 & +77 & +124 & +60 & +155 & +9 & +72 & +123 \\
email\_capped\_8 & +76 & +15 & +41 & +59 & +157 & +50 & +77 & +96 & +60 & +167 & -1 & +73 & +98 \\
\bottomrule
\end{tabular}
}
\end{table*}

\begin{table*}[h]
\centering
\caption{Graduate school discrimination gap (\$/yr) by information condition, averaged over intents and motivations.}
\label{tab:gradschool-info}
\resizebox{\textwidth}{!}{
\begin{tabular}{l*{13}{c}}
\toprule
Condition & GPT-5 & \shortstack{GPT-5-\\mini} & \shortstack{GPT-5-\\nano} & GPT-5.5 & \shortstack{Claude\\Opus 4.8} & \shortstack{Claude\\Sonnet 5} & \shortstack{Claude\\Haiku 4.5} & \shortstack{Gemini 2.5\\Flash} & \shortstack{Gemini 3.1\\Flash Lite} & \shortstack{Gemini 3\\Flash} & \shortstack{Qwen3.5-\\2B} & \shortstack{Qwen3.5-\\9B} & \shortstack{Qwen3.5-\\35B-A3B} \\
\midrule
in\_context & +1,599 & +2,859 & +2,078 & +1,644 & +2,576 & +2,623 & +2,898 & +3,938 & +8,214 & +9,337 & +267 & +3,070 & +3,605 \\
tool\_full & +1,061 & +2,647 & +1,622 & +763 & +3,467 & +2,406 & +2,322 & +607 & +5,591 & +11,081 & -2,056 & +2,872 & +3,827 \\
tool\_blocked\_financial & +117 & +4 & +152 & -114 & -58 & +50 & +27 & -1,001 & -648 & +656 & +1,792 & -471 & -403 \\
tool\_blocked\_employment & +1,678 & +2,501 & +2,558 & +664 & +4,486 & +3,570 & +3,105 & +682 & +6,384 & +11,787 & +2,911 & +3,609 & +4,470 \\
tool\_blocked\_health & +1,733 & +2,279 & +2,341 & +635 & +3,803 & +2,740 & +2,962 & +3,926 & +5,193 & +15,226 & -627 & +3,299 & +3,290 \\
tool\_blocked\_life\_events & +2,542 & +2,887 & +2,636 & +1,228 & +3,993 & +3,550 & +3,128 & +2,427 & +6,226 & +10,513 & -59 & +2,694 & +4,908 \\
tool\_blocked\_demographics & +1,709 & +2,197 & +1,886 & +745 & +3,767 & +2,825 & +2,590 & +2,020 & +6,824 & +18,106 & -163 & +2,360 & +3,545 \\
email\_full & +783 & +1,344 & +385 & +445 & +1,583 & +1,176 & +593 & +3,650 & +295 & +3,467 & -866 & +649 & +2,121 \\
email\_capped\_0 & -121 & +48 & -402 & -59 & +18 & +39 & -197 & +868 & +23,867 & +5 & -137 & +389 & +90 \\
email\_capped\_2 & +3,376 & +1,976 & +13 & +1,389 & +5,589 & +4,800 & +738 & +4,864 & -124 & +4,392 & -359 & +305 & +1,223 \\
email\_capped\_4 & +1,915 & +1,708 & -72 & +1,012 & +3,745 & +2,350 & +1,071 & +4,081 & +7,446 & +3,678 & +1,486 & +275 & +1,623 \\
email\_capped\_6 & +1,516 & +1,522 & -507 & +726 & +2,614 & +1,869 & +408 & +3,227 & +385 & +4,760 & -1,575 & +586 & +1,654 \\
email\_capped\_8 & +871 & +1,718 & +544 & +551 & +1,681 & +1,112 & +1,280 & +2,476 & +128 & +2,492 & +638 & +1,487 & +1,587 \\
\bottomrule
\end{tabular}
}
\end{table*}

\subsection{Blocked-Axis Analysis}

Beyond varying the overall method of access, we want to understand which part of the persona is actually associated with the recommendation gap. We therefore block one axis at a time while leaving the other four available. As shown for five representative models in Table~\ref{tab:blocking}, blocking the Financial axis reduces the flight gap to near zero. This does not happen when Employment, Health, Life Events, or Demographics is blocked, where the gap remains positive.
\begin{table}[t]
\centering
\caption{Flight discrimination gap (\$) by blocked axis, with all intents pooled. Appendix~\ref{app:information-access} reports all models and domains.}
\label{tab:blocking}
\small
\begin{tabular}{lccccc}
\toprule
\textbf{Blocked axis}
  & \textbf{GPT-5} & \shortstack{\textbf{Gemini 2.5}\\\textbf{Flash}}
  & \shortstack{\textbf{Gemini 3}\\\textbf{Flash}} & \textbf{Qwen3.5-9B} & \shortstack{\textbf{Qwen3.5-}\\\textbf{35B-A3B}} \\
\midrule
\textit{None (tool-full)} & \textit{+107} & \textit{+177} & \textit{+145} & \textit{+138} & \textit{+141} \\
\textbf{Financial}        & \textbf{$-$11} & \textbf{+17} & \textbf{$-$4} & \textbf{$-$5} & \textbf{+10}  \\
Employment                & +117 & +197 & +192 & +153 & +170 \\
Health                    & +92  & +155 & +150 & +135 & +118 \\
Life Events               & +113 & +178 & +175 & +136 & +144 \\
Demographics              & +119 & +178 & +172 & +150 & +146 \\
\bottomrule
\end{tabular}
\end{table}

\subsection{Intent-Conditioned Results}

The user's stated intent may also change how persona information affects the recommendation. We test this under the neutral, cheap, quality, and ceiling requests introduced earlier. The resulting intent-conditioned gaps appear in Figure~\ref{fig:intent_override} of the main Results section.

\subsection{Retrieval Versus Recommendation Steering}
\label{app:gpt-5-nano-case-study}

\begin{table}[h]
\centering
\caption{Financial-axis retrieval and recommendation effects for all 13 models in the tool-full condition.}
\label{tab:gpt-5-nano-case-study}
\begin{tabular}{lcccc}
\toprule
Model & Financial Retrieval (\%) & Flights $d$ & Insurance $d$ & Flights Gap (\$) \\
\midrule
GPT-5 & 100.0 & 0.41 & 0.42 & +107 \\
GPT-5-mini & 97.9 & 0.29 & 0.29 & +74 \\
GPT-5-nano & 87.1 & 0.07 & 0.15 & +13 \\
GPT-5.5 & 100.0 & 0.39 & 0.28 & +92 \\
\midrule
Claude Opus 4.8 & 100.0 & 1.01 & 0.73 & +198 \\
Claude Sonnet 5 & 100.0 & 0.83 & 0.33 & +176 \\
Claude Haiku 4.5 & 100.0 & 0.83 & 0.38 & +141 \\
\midrule
Gemini 2.5 Flash & 93.0 & 0.76 & 0.48 & +177 \\
Gemini 3.1 Flash Lite & 99.4 & 0.44 & 0.61 & +112 \\
Gemini 3 Flash & 100.0 & 0.56 & 0.84 & +145 \\
\midrule
Qwen3.5-2B & 30.2 & 0.10 & 0.07 & +14 \\
Qwen3.5-9B & 95.3 & 0.71 & 0.46 & +138 \\
Qwen3.5-35B-A3B & 95.8 & 0.62 & 0.30 & +141 \\
\bottomrule
\end{tabular}
\end{table}

Finally, accessing personal information does not necessarily mean that the model will use it to steer its recommendations. The GPT-5-nano example above is the most starkly different: it retrieves the Financial axis in 87.1\% of tool-full trials but has a flights effect of just $d=0.07$, while Qwen3.5-9B retrieves 95.3\% of the axes but has a flights effect of $d=0.71$; in other words, the two models have very similar retrieval rates despite very different flights effects. Retrieval and steering direction are therefore not perfectly correlated. However, in this study, we do not go into detail on why this behavior may be occurring and only describe it.

\section{Reproducibility and Ethics}

In this work, we aim to study and scope the behavior of agents while making economic recommendations across a multitude of domains. We believe that studying these behaviors will allow us to examine an emerging axis of AI safety that may become increasingly important as these models are more widely deployed in personalization tasks.

\subsection{Artifact and Release Information}

To ensure the replicability of our study for future evaluations and safety audits, we open-source our harness, environments, MCP servers, access protocols, MCPs, and the agent prompts themselves. To extend the study, we also release the persona-generation, inventory-generation, and email-synthesis prompts alongside the aforementioned materials. We release our runners for the OpenAI, Anthropic, and Gemini APIs, as well as our vLLM inference configuration. We hope this will allow the broader community to replicate our findings, test additional and newer models, vary tool configurations (e.g., removing persona tools entirely while retaining search), and test alternative prompt framings.

\subsection{Ethics Statement}

Our work aims to provide an analytical lens into model behavior and personalized recommendations. To ensure consistency within our scope and prevent the leakage of actual personal information, we designed our experiments around synthetic user data. However, we ensure that our observations capture both the variation caused by access to this data and the behavior of actively constructing user profiles and accessing this kind of information before providing product recommendations.

All personas, emails, and inventories that we've used in our experimentation have been synthetically-generated. There were no human users and no personal information gathered. The experiments consist of API calls to open-weight or public APIs or models.

\newpage

\end{document}